\documentclass[journal]{IEEEtran}

\ifCLASSOPTIONcompsoc
  \usepackage[nocompress]{cite}
\else
  \usepackage{cite}
\fi

\usepackage{multirow}
\usepackage{gensymb}
\usepackage{colortbl}
\usepackage{nicefrac}
\definecolor{tabfirst}{rgb}{1, 0.7, 0.7} % red
\definecolor{tabsecond}{rgb}{1, 0.85, 0.7} % orange
\definecolor{tabthird}{rgb}{1, 1, 0.7} % yellow

\ifCLASSINFOpdf
\else
\fi
\usepackage{graphicx}
\usepackage{amsmath}
\usepackage{amssymb}
\usepackage{booktabs}
\usepackage{bbding}
\usepackage{color,xcolor}
\usepackage[svgnames, table]{xcolor}
\usepackage{multirow}
\usepackage{float}
\usepackage{adjustbox}
\usepackage{url}
\usepackage{diagbox}
\usepackage[switch]{lineno}
\usepackage[hidelinks]{hyperref}
\usepackage[capitalize]{cleveref}
\usepackage{soul}

\newcommand{\modify}[1]{\textcolor{black}{#1}}

\begin{document}
% \linenumbers

\newcommand{\reflabel}{dummy} % Dummy initial reflabel - use renewcommand...

% \newcommand{\be}{\begin{equation}}
% \newcommand{\ee}{\end{equation}}
% \newcommand{\eqlabel}[1]{\label{eq:\reflabel-#1}}
% \renewcommand{\eqref}[2][\reflabel]{(\ref{eq:#1-#2})}

% Generic reference commands
\newcommand{\seclabel}[1]{\label{sec:\reflabel-#1}}
\newcommand{\secref}[2][\reflabel]{Section~\ref{sec:#1-#2}}
\newcommand{\Secref}[2][\reflabel]{Section~\ref{sec:#1-#2}}
\newcommand{\secrefs}[3][\reflabel]{Sections~\ref{sec:#1-#2} and~\ref{sec:#1-#3}}

\newcommand{\eqlabel}[1]{\label{eq:\reflabel-#1}}
\renewcommand{\eqref}[2][\reflabel]{(\ref{eq:#1-#2})}
\newcommand{\Eqref}[2][\reflabel]{(\ref{eq:#1-#2})}
\newcommand{\eqrefs}[3][\reflabel]{(\ref{eq:#1-#2}) and~(\ref{eq:#1-#3})}

\newcommand{\figlabel}[2][\reflabel]{\label{fig:#1-#2}}
\newcommand{\figref}[2][\reflabel]{Fig.~\ref{fig:#1-#2}}
\newcommand{\Figref}[2][\reflabel]{Fig.~\ref{fig:#1-#2}}
\newcommand{\figsref}[3][\reflabel]{Figs.~\ref{fig:#1-#2} and~\ref{fig:#1-#3}}
\newcommand{\Figsref}[3][\reflabel]{Figs.~\ref{fig:#1-#2} and~\ref{fig:#1-#3}}

\newcommand{\tablelabel}[2][\reflabel]{\label{table:#1-#2}}
\newcommand{\tableref}[2][\reflabel]{Table~\ref{table:#1-#2}}
\newcommand{\Tableref}[2][\reflabel]{Table~\ref{table:#1-#2}}
\newcommand{\etal}{\emph{et al.}}
\newcommand{\eg}{e.g.}
\newcommand{\ie}{i.e. }
\newcommand{\etc}{etc. }

%%%
%%% Stuff for bold maths typesetting  ----------------------------------------
%%%
%%%   e.g. use "\bfmu" for boldface mu symbol
%%%
\def\bfmu{\mbox{\boldmath$\mu$}}
\def\bftau{\mbox{\boldmath$\tau$}}
\def\bftheta{\mbox{\boldmath$\theta$}}
\def\bfdelta{\mbox{\boldmath$\delta$}}
\def\bfphi{\mbox{\boldmath$\phi$}}
\def\bfpsi{\mbox{\boldmath$\psi$}}
\def\bfeta{\mbox{\boldmath$\eta$}}
\def\bfnabla{\mbox{\boldmath$\nabla$}}
\def\bfGamma{\mbox{\boldmath$\Gamma$}}

%%% Make figure placement a little more predictable.
% We trust the user to move figures if this results
% in ugliness.
% Minimize bad page breaks at figures
%\renewcommand{\textfraction}{0.01}
%\renewcommand{\floatpagefraction}{0.99}
%\renewcommand{\topfraction}{0.99}
%\renewcommand{\bottomfraction}{0.99}
%\renewcommand{\dblfloatpagefraction}{0.99}
%\renewcommand{\dbltopfraction}{0.99}
%\setcounter{totalnumber}{99}
%\setcounter{topnumber}{99}
%\setcounter{bottomnumber}{99}
%
%% Add a period to the end of an abbreviation unless there's one
%% already, then \xspace.
%\makeatletter
%\DeclareRobustCommand\onedot{\futurelet\@let@token\@onedot}
%\def\@onedot{\ifx\@let@token.\else.\null\fi\xspace}
%
%\def\eg{\emph{e.g}\onedot} \def\Eg{\emph{E.g}\onedot}
%\def\ie{\emph{i.e}\onedot} \def\Ie{\emph{I.e}\onedot}
%\def\cf{\emph{c.f}\onedot} \def\Cf{\emph{C.f}\onedot}
%\def\etc{\emph{etc}\onedot} \def\vs{\emph{vs}\onedot}
%\def\wrt{w.r.t\onedot} \def\dof{d.o.f\onedot}
%\def\etal{\emph{et al}\onedot}
%\makeatother

% ---------------------------------------------------------------

\newcommand{\R}{\mathbb{R}}

\definecolor{tabfirst}{rgb}{1, 0.7, 0.7} % red
\definecolor{tabsecond}{rgb}{1, 0.85, 0.7} % orange
\definecolor{tabthird}{rgb}{1, 1, 0.7} % yellow

\title{Improving Local Feature Matching by Entropy-inspired Scale Adaptability and Flow-endowed Local Consistency}

\author{Ke Jin,  Jiming Chen, \IEEEmembership{Fellow, IEEE}, and Qi Ye, \IEEEmembership{Member, IEEE}
\IEEEcompsocitemizethanks{\IEEEcompsocthanksitem Kejin, Jiming Chen, and Qi Ye are with the State Key Laboratory of Industrial Control Technology, Zhejiang University. Email: \{kejin27, cjm, qi.ye\}@zju.edu.cn.
Corresponding author: \emph{Qi Ye}. 

This work was supported by NSFC under Grants 62088101. 
}}
% <-this % stops an unwanted space
% \thanks{Manuscript received April 19, 2005; revised August 26, 2015.}

\markboth{IEEE Transactions on Circuits and Systems for Video Technology, 2026}%
{}

\IEEEpubid{%
\parbox{1.8\columnwidth}{\centering
Copyright © 2026 IEEE. Personal use of this material is permitted. Permission from IEEE
must be obtained for all other uses. This is the author's accepted version of
the article. The final version of record is available at DOI:
10.1109/TCSVT.2026.3681288
}}

\maketitle

\begin{figure*}[!t]
    \centering
    \includegraphics[width=0.95\linewidth]{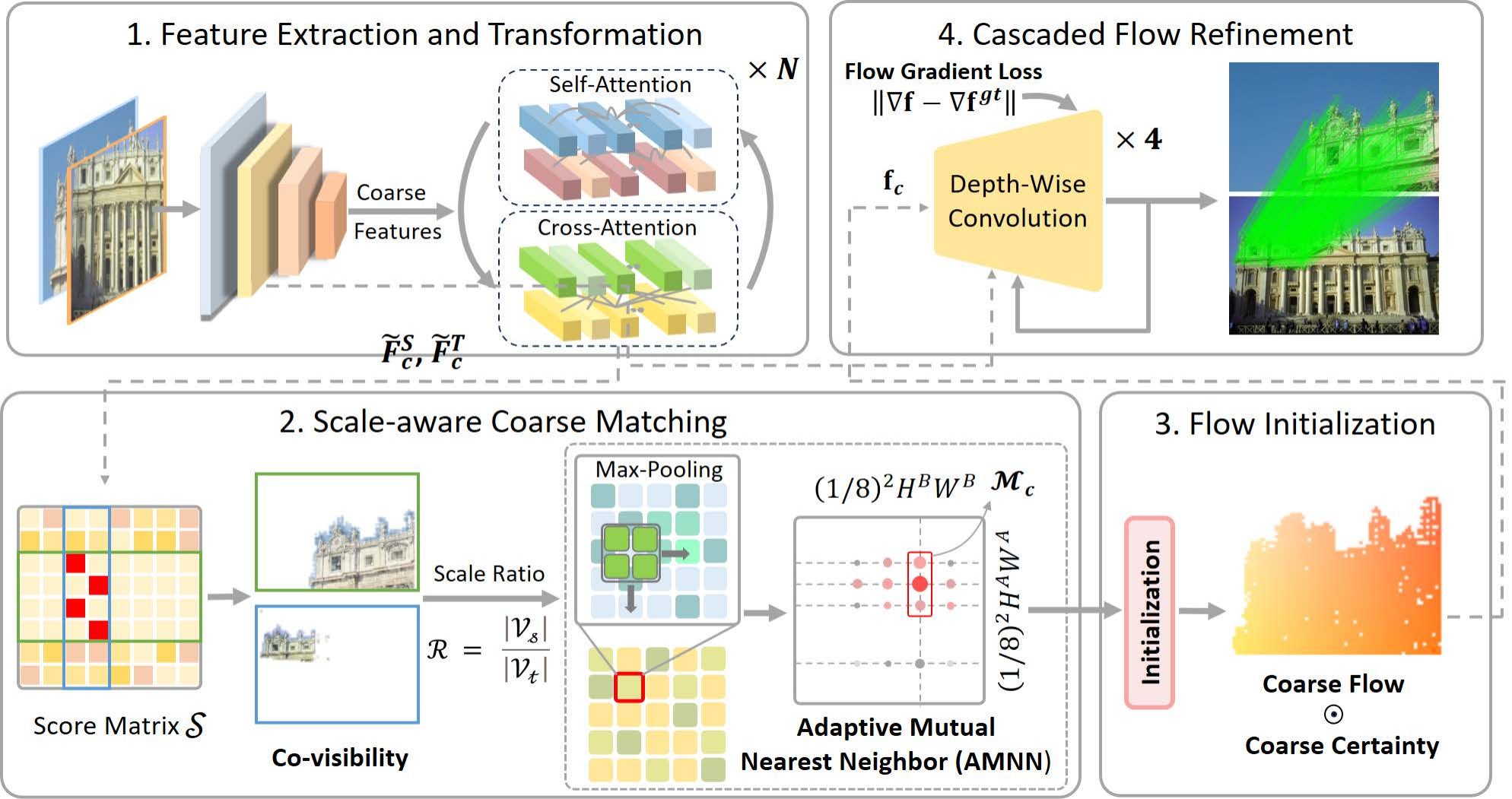}%
    \caption{\textbf{Overview of the proposed method.}
    \textbf{(1)} Given a pair of images, a standard Siamese CNN extracts multi-level features, and the coarsest features $F_{c}^{S}$ and $F_{c}^{T}$ are transformed to be more discriminative after processed by interleaved self and cross attention layers.
    \textbf{(2)} We approximate the scale ratio based on the entropy of the score matrix $\mathcal{S}$ which is computed by correlation between the transformed features $\Tilde{F}_{c}^{S}$ and $\Tilde{F}_{c}^{T}$. Coarse matches $\{\mathcal{M}_{c}\}$ are then obtained according to the adaptive mutual nearest neighbor (AMNN) criterion.
    \textbf{(3)} Given established coarse matches, the coarse flow and certainty are initialized in a straightforward way. 
    \textbf{(4)} A cascade of depth-wise convolution modules are employed to refine the flow progressively, yielding the final pixel-level correspondences.
    }
    \label{fig:overview}
    \vspace{-0.3cm}  
\end{figure*}

\begin{abstract}
Recent semi-dense image matching methods have achieved remarkable success, but two long-standing issues still impair their performance. At the coarse stage, the over-exclusion issue of their mutual nearest neighbor (MNN) matching layer makes them struggle to handle cases with scale difference between images. 
To this end, we comprehensively revisit the matching mechanism and make a key observation that the hint concealed in the score matrix can be exploited to indicate the scale ratio. Based on this, we propose a scale-aware matching module which is exceptionally effective but introduces negligible overhead.
At the fine stage, we point out that existing methods neglect the local consistency of final matches, which undermines their robustness. To this end, rather than independently predicting the correspondence for each source pixel, we reformulate the fine stage as a cascaded flow refinement problem and introduce a novel gradient loss to encourage local consistency of the flow field.
Extensive experiments demonstrate that our novel matching pipeline, with these proposed modifications, achieves robust and accurate matching performance on downstream tasks.
\end{abstract}


\section{Introduction}\label{sec:introduction}
% \yq{need to write out the full word of mmWave}
\IEEEPARstart{I}{mage} matching, which aims to find a set of highly accurate correspondences given an image pair, is the cornerstone of 3D vision techniques~\cite{Agarwal2009BuildingRI,schonbergerStructurefromMotionRevisited2016,Lindenberger2021PixelPerfectSW,he2023detectorfree} and can facilitate many computer graphics tasks including novel view synthesis (NVS)~\cite{sparf2023, lao2023corresnerf, bösiger2024marinerenhancingnovelviews} and animation~\cite{corrAnimation}.
Recent image matching methods can be generally classified into three categories: sparse, semi-dense, and dense methods. Among them, the semi-dense methods~\cite{sun2021loftr, wang2022matchformer, chen2022aspanformer, tang2022quadtree, yu2023adaptive, giang2022topicfm, wang2024efficient} overcome the limitations of sparse methods~\cite{luo2020aslfeat,tian2020d2d,sarlin20superglue,lindenberger2023lightglue} in challenging scenarios, such as poor texture and repetitive patterns, while tending to be more efficient than their dense counterparts~\cite{edstedt2023dkm,edstedt2023roma}.
They capture long-range dependencies with transformers and establish correspondences in a coarse-to-fine paradigm. Specifically, they first employ the dual-softmax operator as a matching layer and determine coarse matches according to the mutual nearest neighbor (MNN) criterion. Then the feature locations of coarse matches on one image are fixed, while their sub-pixel correspondences are refined based on the feature correlation with the coarse-matched feature patch cropped from the other image.

However, as some previous work points out~\cite{Ni2023PATS, huang2023adaptive, wang2024efficient}, the existing semi-dense methods still suffer from two long-standing issues, which hinder their performance in real-world applications. 
At the coarse stage of semi-dense methods, the MNN matching criterion implemented by a dual-softmax layer tends to over-exclude plausible coarse matches~\cite{Ni2023PATS, huang2023adaptive}. When scale variation between image pairs exists, the one-to-one matching constraint inevitably leads to degraded performance.
During the sub-pixel refinement phase, although existing methods introduce a cascaded refinement strategy to improve per-pixel accuracy, the minor errors cannot be completely eliminated due to the presence of the spatial-variance problem~\cite{wang2024efficient}. Since all source pixels are refined independently, the predicted offsets deviating from the ground truth are random, leading to local inconsistency in the final matching results.

In this paper, we revisit the semi-dense image matching pipeline by tackling the over-exclusion and local inconsistency issues, yet maintaining the high efficiency which is critical for many downstream tasks like visual odometry. 
Previous work~\cite{Ni2023PATS, huang2023adaptive} addresses the over-exclusion issue by introducing co-visible area segmentation~\cite{huang2023adaptive} or iterative patch subdivision~\cite{Ni2023PATS} techniques, but makes the pipeline cumbersome and time-consuming.
In contrast, we dig into the matching mechanism and propose a simple yet exceptionally effective solution to this issue. A key observation that motivates us is that the entropy of the score matrix, an intermediate outcome in the coarse stage, inherently contains heuristic information indicating the co-visibility and the scale variation.
Based on the inferred scale, an adaptive inspection mechanism is introduced to the reverse check of the MNN criterion, termed Adaptive MNN (AMNN). After comprehensively investigating the design of the existing matching layer (dual-softmax), we integrate the AMNN criterion into it smoothly, bringing in negligible computational cost and maintaining the differentiable property.

\IEEEpubidadjcol

To improve the local consistency of the final results from the fine stage, our insight is to leverage the surrounding context of each source pixel during refinement rather than solely relying on the correlation between each source pixel and its corresponding target patch.
Specifically, we replace the existing fine stage with cascaded flow refinement, which leverages convolutional modules to aggregate the neighboring hints of the flow field. 
To explicitly enhance the local consistency of the flow results, we further introduce a novel loss for refinement, which supervises the gradients of the flow field to encourage the continuity and smoothness of neighboring pixels in the flow field. 
As an additional benefit from the cascaded flow refinement, our method ultimately yields dense correspondences, which is beneficial for downstream tasks, within the semi-dense framework.

Extensive experiments on various benchmarks demonstrate that our method significantly benefits from these modifications, which are based on a comprehensive review of existing semi-dense methods. More ablation studies also show that the novel matching pipeline performs robustly against factors such as image resolution and scale variance compared to previous semi-dense methods.

In summary, this paper has the following contributions:

(1) We propose a novel matching pipeline with multiple improvements to existing semi-dense methods, achieving competitive performance on various benchmarks.

(2) We propose a scale-aware coarse matching mechanism to address the over-exclusion issue, which is simple but exceptionally effective.

(3) We propose to utilize cascaded flow refinement with a novel gradient loss for sub-pixel refinement, tackling the local inconsistency issue in the fine stage of existing semi-dense methods.

\begin{figure*}[!ht]
    \centering
    \includegraphics[width=\textwidth]{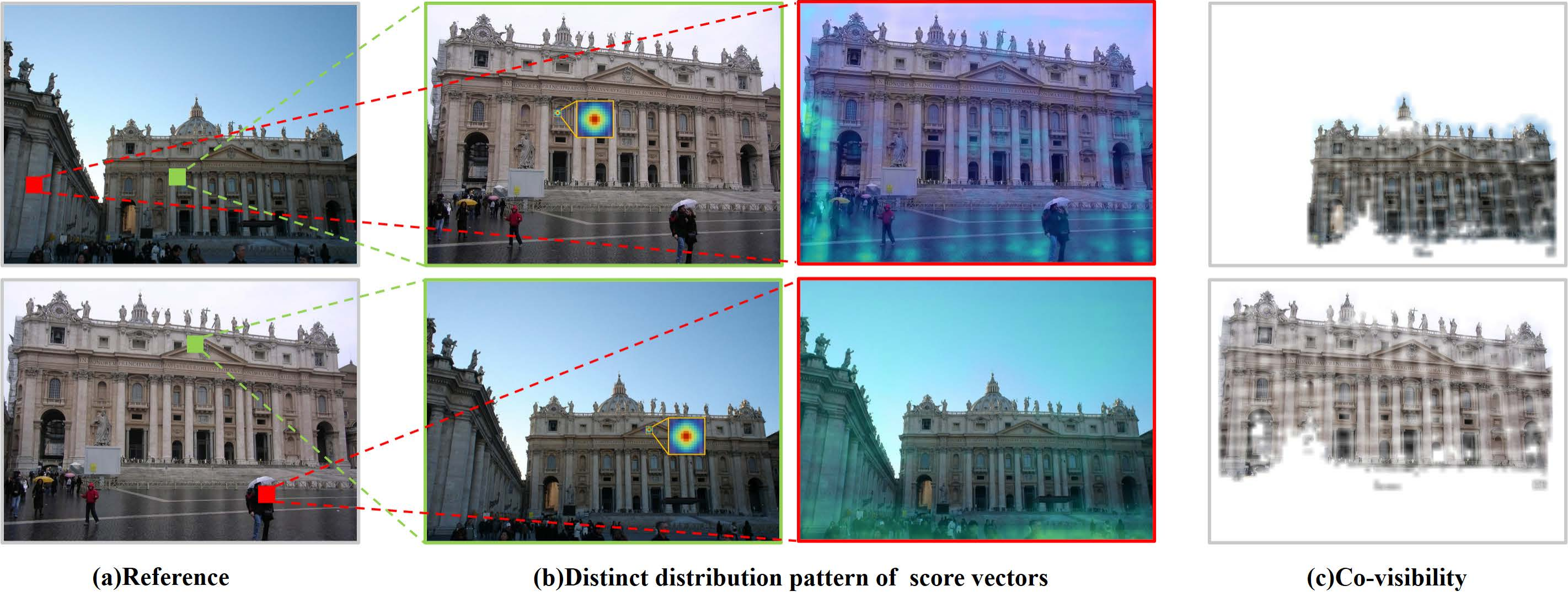}
    \vspace{-0.7cm}
    \caption{
    \textbf{The visualization of matching score heatmaps and inferred co-visibility.} We sample four points on the reference image pair (a), where the red markers indicate unmatchable points and the green markers indicate co-visible points. (b) shows their matching score heatmaps with the other image, which exhibit completely different distribution patterns: one-hot and evenly distributed. Motivated by this, the entropy of score vectors can be used to infer the co-visible regions of each image (c), enabling scale-aware matching.
    }
    \label{fig:covisibility}
\end{figure*}

\section{Related Work}
\label{sec:related_work}

\subsection{NN Search Based Image Matching.}
Traditional methods~\cite{LoweDavid2004DistinctiveIF, rosten2006machine, 
bay2008speeded,ORB, rosten2006machine} detect keypoints, describe local regions, and match them via nearest-neighbor search.  Many elaborately designed descriptors~\cite{savinov2017quad, barroso2019key, tian2017l2, mishchuk2017working, tian2019sosnet, ebel2019beyond,dusmanu2019d2,DeTone2017SuperPointSI,revaud2019r2d2,luo2020aslfeat,tian2020d2d, RWBD, Regress_Desc, TCDesc, Scale_Desc} are then introduced to enhance their discriminability and robustness.

SuperGlue~\cite{sarlin20superglue} is the first to introduce attention mechanisms to image matching and formulates the matching task as a partial assignment between two sets. LightGlue~\cite{lindenberger2023lightglue}
adaptively prunes attention to reduce the inference overhead of ~\cite{sarlin20superglue} while maintaining competitive performance.

Semi-dense matching methods~\cite{sun2021loftr, wang2022matchformer, chen2022aspanformer, tang2022quadtree, yu2023adaptive, giang2022topicfm, wang2024efficient} further eliminate the dependency on keypoints and implement mutual nearest neighbor (MNN) matching on each patch of the image, followed by a pixel-level refinement. LoFTR~\cite{sun2021loftr} first introduces this matching paradigm and achieves revolutionary matching performance. Following variants~\cite{wang2022matchformer, chen2022aspanformer, tang2022quadtree, giang2022topicfm, wang2024efficient, MSFA-Net, ContextMatcher} further improve accuracy and inference speed of LoFTR~\cite{sun2021loftr} through meticulously designed attention mechanisms.

\subsection{Regression Based Image Matching.}
Dense matching methods~\cite{melekhov2019dgc, truong2021learning,edstedt2023dkm,edstedt2023roma} are designed to estimate all possible correspondences between two images. ~\cite{melekhov2019dgc, truong2021learning} predict the warp by regression based on the global 4D-correlation volume and estimate the presence of a pixel correspondence between the images. DKM~\cite{edstedt2023dkm} employs a kernel regression global matcher for coarse regression and refines the warp through stacked feature maps and depthwise convolutions. ROMA~\cite{edstedt2023roma} leverages frozen pretrained features from the foundation model DINOv2~\cite{oquab2023dinov2} to match under extreme view changes. 

These methods~\cite{edstedt2023dkm,edstedt2023roma} achieve excellent performance and robustness, but are computationally expensive. 
In comparison, our matching pipeline can also obtain dense results while maintaining the efficient framework of semi-dense methods.

\subsection{Semi-dense methods designed for scale variance}
Similar to us, some previous methods~\cite{Ni2023PATS, huang2023adaptive} attempt to address the over-exclusion problem caused by scale variance.
PATS~\cite{Ni2023PATS} formulates patch matching as a patch area transportation problem and adopts a patch subdivision strategy that iteratively crops the patches and resizes the image content to align the scale between images to alleviate appearance distortion. Although such pipeline achieves competitive performance, it leads to extremely high computational cost.
AdaMatcher ~\cite{huang2023adaptive} performs scale analysis by introducing a co-visible area segmentation stage which incurs extra computational cost.
Then it performs standard MNN matching on features from different levels ($\frac{1}{8}$, $\frac{1}{4}$, or $\frac{1}{2}$) according to the inferred scale, which lacks flexibility and increases the burden on the matching layer.
In contrast, our method avoids the introduction of additional modules and maintains the conciseness of the LoFTR pipeline~\cite{sun2021loftr} while effectively addressing the problem of over-exclusion.

\section{Method}
\label{sec:method}
Given a source image $\*I_{S}$ and a target image $ \*I_{T}$, our method aims to estimate all possible correspondences between them.
We first establish coarse matching via a scale-aware coarse matching module, which is used to initialize a reliable coarse flow map and a certainty map. Then we refine the flow map in a cascaded manner and select the final pixel-level correspondences based on the certainty. An overview of our method is shown in Fig.~\ref{fig:overview}.

\subsection{Feature Extraction and Transformation}
\label{subsec:feature}
Following common practice in semi-dense image matching methods, we adopt a Siamese CNN to extract features. Then a transformer is used to enhance the discriminability of the features at the coarse level.

% 2. Feature extraction
A standard CNN architecture extracts multi-level features from the two images at $\frac{1}{8}$, $\frac{1}{4}$, and $\frac{1}{2}$ of the original resolutions. We denote the coarsest features as $F_{c}^{S}$ and $F_{c}^{T}$. 
Benefiting from the inductive bias and translational invariance properties of CNNs, these features capture rich information about local textures and geometric details.
The coarse features $F_{c}^{S}$ and $F_{c}^{T}$ are row-wise flattened into vectors and processed through a transformer module, which alternately stacks self-attention and cross-attention layers. This transformation enables the features to more effectively capture long-range dependencies, ultimately improving the accuracy of the matching process. We denote the transformed features as $\Tilde{F}_{c}^{S}$ and $\Tilde{F}_{c}^{T}$ .

\subsection{Scale-aware Coarse Matching}
\label{subsec:coarse}
% 1. Section overview
Given the transformed features as $\Tilde{F}_{c}^{S}$ and $\Tilde{F}_{c}^{T}$, we aim to identify all plausible correspondences between the two sets of patches. As discussed above, LoFTR~\cite{sun2021loftr} and its variants~\cite{wang2022matchformer, chen2022aspanformer, tang2022quadtree, giang2022topicfm, wang2024efficient} adopt the mutual nearest neighbor (MNN) criterion, which suffers from the over-exclusion issue caused by scale variance.
To address this issue and design a scale-aware coarse matching module, we derive heuristic information concealed in the score matrix to approximate the scale, and introduce an adaptive matching mechanism which is simple but exceptionally effective.

% \begin{color}{red}
\paragraph{Preliminaries.} Following prior semi-dense detector-free pipelines such as LoFTR~\cite{sun2021loftr} and EfficientLoFTR~\cite{wang2024efficient}, coarse matching is performed on the transformed coarse features through a differentiable matching layer. Existing methods typically construct a score matrix $\mathcal{S}$ between the transformed features by $\mathcal{S}(i,j)=\frac{1}{\tau} \cdot \langle \Tilde{F}_{c}^{A}(i), \Tilde{F}_{c}^{B}(j) \rangle$, where $\tau$ is a temperature parameter, and then apply the dual-softmax operator~\cite{Rocco2018NeighbourhoodCN,tyszkiewicz2020disk,sun2021loftr} on both dimensions of $\mathcal{S}$ to obtain the probability of soft mutual nearest neighbor matching. Specifically, the matching probability $P_{c}$ is obtained by:
\begin{equation}
    \mathcal{P}_{c}(i,j)=\operatorname{softmax}(\mathcal{S}(i,\cdot))_{j} \cdot \operatorname{softmax}(\mathcal{S}(\cdot,j))_{i}.
\end{equation}
% \end{color}

According to the MNN criterion, a patch pair $(i,j)$ is considered matched only when $P_{c}(i,j)$ is not only the column maximum but also the row maximum, which is designed to reject unmatchable pairs caused by occlusion or non-repeatability. However, this criterion enforces an overly strict constraint that only one-to-one matching is allowed.
When scale variance exists between the image pairs, existing methods tend to struggle since they are not able to handle the many-to-one cases.

\paragraph{Scale Approximation.}
Previous work introduces extra cumbersome pipelines to adapt to the unknown scale variance, which we find unnecessary after comprehensive investigation.
The key observation is that the score matrix $\mathcal{S}$ inherently contains concealed clues about the co-visibility and the scale ratio between image pairs.

% \begin{color}{red}
Consider a coarse patch $i$ in the source image as an example. We denote its row-wise normalized matching score distribution over all target patches as
\begin{equation}
    p_i(j)=\operatorname{softmax}(\mathcal{S}(i,\cdot))_j,
\end{equation}
where $\sum_j p_i(j)=1$.
Since cross entropy loss is adopted to supervise the matching probability $P_{c}(i,j)$, it implicitly drives the score distributions toward two distinct regimes, as shown in Fig.~\ref{fig:covisibility}.
Specifically, when the patch $i$ represents a co-visible region between image pairs, the normalized distribution tends to be concentrated (one-hot like). When the patch $i$ is unmatchable, the scores tend to be more evenly distributed.

Based on this observation, we use Shannon entropy as a co-visibility indicator for patch $i$: 
\begin{equation}
    e(i) = -\sum_{j} p_i(j) \cdot \log\big(p_i(j)\big).
\end{equation}
Low entropy implies a concentrated matching distribution (co-visible), whereas high entropy implies a diffuse distribution (unmatchable).
Then, the set of co-visible patches in the source image is estimated by:
\begin{equation}
    \mathcal{V}_{s} = \{i \mid e(i) < \theta_{e}\}
    \label{eq:threshold}
\end{equation}
Similarly, we can obtain the set of co-visible patches in the target image $\mathcal{V}_{t}$. Then the scale ratio between the image pair can be approximated by:
\begin{equation}
    \mathcal{R} = \frac{|\mathcal{V}_{s}|}{|\mathcal{V}_{t}|}
\end{equation}
where $|\cdot|$ denotes set cardinality. Here, $\mathcal{R}$ provides a coarse global cue for adapting the AMNN inspection window.
% \end{color}

\begin{figure}[!tb]
    \centering
    \includegraphics[width=0.85\linewidth]{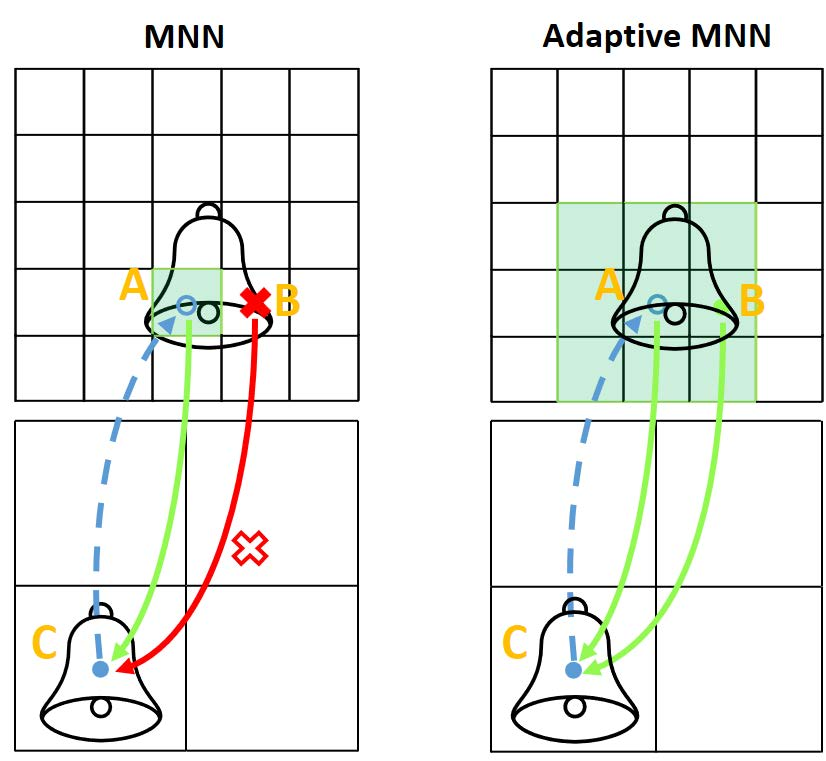}
    \caption{\textbf{Illustration for inspection window.}
    The many-to-one case is used to illustrate the benefits of the inspection window introduced by AMNN. For better visualization, the source image in the first row and the target image in the second row are shown in different scales. \modify{The green translucent square denotes the inspection window.} The solid arrows to the target image indicate the predicted nearest neighbors for patches in the source image, while the dashed arrows in reverse direction indicate the cycle consistency check after that. \modify{In this example, AMNN produces a scale-dependent window with size $3$, which enables the correspondence between B and C to be accepted.}}
    \label{fig:AMNN}
\end{figure}

\paragraph{Adaptive MNN Matching.}
Based on the approximated scale ratio, we can design an adaptive MNN matching criterion.
Specifically, we introduce an inspection window to relax the reverse check on the image with larger scale, as shown in Fig.~\ref{fig:AMNN}.
\modify{The green translucent square is the inspection window, and its scale-dependent size is $3$ in this example.}
Assuming the scale of the source image is larger, the set of valid matches for AMNN can be defined as:
\begin{equation}
    \{(i, j)| \operatorname{NN}^{S \rightarrow T}(p_{i}^{S}) = p_{j}^{T}, \operatorname{NN}^{T \rightarrow S}(p_{j}^{T}) \in U(p_{i}^{S}, \delta) \},
\end{equation}
where we denote the coordinate of $i$-th patch in the coarse feature map $\Tilde{F}_{c}^{k}$ as $p_{i}^{k}$, and the coordinate of its nearest neighbor in the other image as $\operatorname{NN}(p_{i}^{k})$. $U(p, \delta)$ corresponds to the inspection window centered at $p_{i}^{S}$ with size $\sqrt{\mathcal{R}}$.

According to this criterion, adaptive many-to-one matching is allowed while the ability to reject mismatches is also preserved.

To accommodate the proposed AMNN criterion, a simple but exceptionally effective modification is done to the standard dual-softmax operator:
\begin{equation}
    \mathcal{P}_{c}(i,j)=\operatorname{softmax}(\mathcal{S}(i,\cdot))_{j} \cdot \phi(\operatorname{softmax}(\mathcal{S}(\cdot,j))_{i}),
\end{equation}
where $\phi$ denotes a max-pooling operation with 
kernel size corresponding to the inspection window size. 
Compared to the standard dual-softmax, the additional max-pooling operation allows $P_{c}(i,j)$ to be activated as long as patch $i$ has a nearest neighbor of patch $j$ within its surrounding, which perfectly fits the idea of AMNN matching.

\subsection{Cascaded Flow Refinement}
\begin{figure}[!tb]
    \centering
    \includegraphics[width=\linewidth]{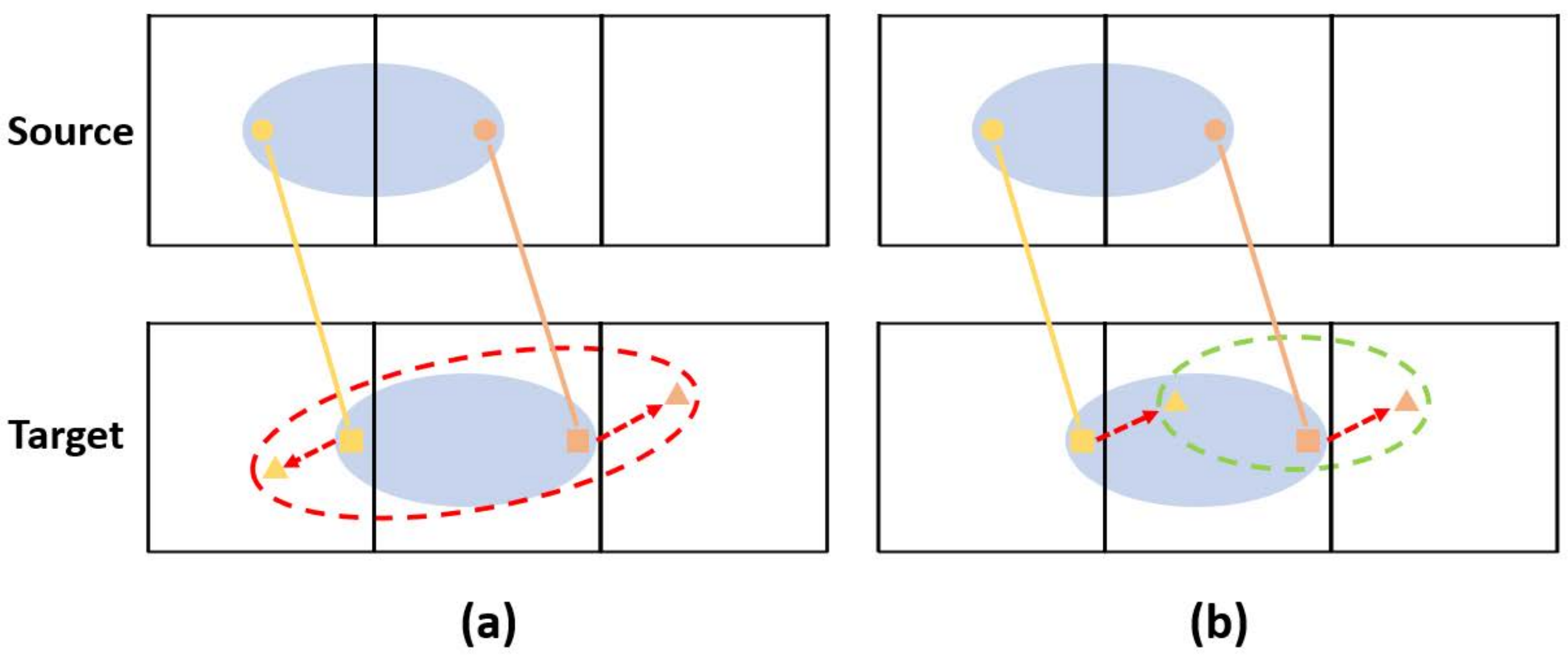}
    \vspace{-0.8cm}
    \caption{\textbf{Illustration for local consistency.} \textbf{(a)} When the predictions of two adjacent source pixels deviate from the ground truth in opposite directions, this can lead to unreasonable distortions in the flow field. \textbf{(b}) While the per-pixel error remains identical to (a), this enhanced local consistency proves beneficial for subsequent downstream tasks.
    }
    \label{fig:Consistency}
\end{figure}

As discussed in~\cite{wang2024efficient}, existing semi-dense methods suffer from spatial-variance problem in the fine stage. Since the pixel features within a target patch are highly similar and difficult to discriminate, the final minor prediction error for each source pixel is inevitable. 
Each source pixel's prediction error is independent, and the error direction is random.
When two adjacent source pixels have predictions that deviate from the ground truth in opposite directions (e.g. one left and one right), it leads to local inconsistency, as illustrated in Fig.~\ref{fig:Consistency}(a).
        
To this end, we propose to replace the existing fine stage with cascaded flow refinement, where the local contextual cues are incorporated by convolutional modules.
We further introduce a novel gradient loss for refinement which enhances the local consistency by encouraging the continuity and smoothness of neighboring pixels in the flow field.
Formally, the objective of our fine stage is to establish 1) a refined flow $\textbf{f}\in \mathbb{R}^{h \times w \times 2}$, indicating the corresponding position in the target image of each pixel from the source image, and 2) a certainty $\mathcal{C}\in \mathbb{R}^{h \times w \times 1}$, representing whether each pixel in the source image is visible in the target image.

\paragraph{Initialization and Refinement.}
Based on the matching probability $\mathcal{P}_{c}$, the coarse flow and certainty can be initialized in a straightforward way. To get the coarse flow $\textbf{f}_{c}\in \mathbb{R}^{h_{c} \times w_{c} \times 2}$, we compute the flow value for each patch $i$ by selecting the index $j$ that maximizes $\mathcal{P}_{c}(i,j)$. Specifically, the flow can be obtained by:
\begin{equation}
    \begin{cases} \textbf{f}_{c}(i)_{x}=\mathop{\arg \max}_{j}\mathcal{P}_{c}(i,j) \pmod{w_{c}} \\
    \textbf{f}_{c}(i)_{y}=\left\lfloor \mathop{\arg \max}_{j}\mathcal{P}_{c}(i,j) \div w_{c} \right\rfloor, \end{cases}
\end{equation}
where $\textbf{f}_{c}(i)$ is the flattened form of $\textbf{f}_{c}(u,v)$,
that $i = u+w_{c}\cdot v$, with $u \in \{0,1,...,w_{c}-1\}$ and $v \in \{0,1,...,h_{c}-1\}$.
The subscript $\{x, y\}$ of $\textbf{f}_{c}(i)$ denotes the offset in the horizontal and vertical directions, respectively.

The initialized flow is then fed into a cascaded convolutional module $R_{\theta}$. The refiner can be decomposed recursively as:
\begin{equation}
    \textbf{f}_{l}=R_{\theta,l}(F_{l}^{S}, F_{l}^{T}, \textbf{f}_{l+1}),
\end{equation}
where $\textbf{f}_{l}$ denotes the flow whose resolution is $\nicefrac{1}{2^{l}}$ of the original image, $ l\in \{3,2,1,0\}$ and $F_{l}^{S}$, $F_{l}^{T}$ are the features at level $l$ of the source and target images, respectively. $R_{\theta,l}$ predicts a residual offset based on the flow from the previous level $\textbf{f}_{l+1}$. 

As for the certainty $\mathcal{C}$, we initialize the coarse certainty from the coarse-level match prediction as:
\begin{equation}
    \mathcal{C}_{c}(i)=\begin{cases} \mathcal{P}_{c}(i,j), & \text{if } \exists \;j, (i,j) \in \mathcal{M}_{c} \\ 0, & \text{otherwise} \end{cases}
\end{equation}
where $\mathcal{M}_{c}$ denotes the set of matched pairs following LoFTR~\cite{sun2021loftr}. After that, we directly upsample $\mathcal{C}_{c}$ using bilinear interpolation to the dense certainty $\mathcal{C} \in \mathbb{R}^{h \times w \times 1}$ which is used to select reliable final pixel matches. 

\paragraph{Flow Gradient Loss}
Previous flow-regression based methods~\cite{edstedt2023dkm,edstedt2023roma} adopt the $l_{2}$ distance between the predicted and ground truth flow as the regression loss for the cascaded refinement, which can be formulated as:
\begin{equation}
    \mathcal{L}_{r} = \sum_{l}\sum_{u,v} \ \left\|\textbf{f}_{l}(u,v) - \textbf{f}_{l}^{gt}(u,v)\right\|_{2}.
\end{equation}

However, we find that this loss is insufficient for our pipeline.
Since AMNN matching is designed to match two sets of patch centers, the coarse flow $\textbf{f}$ is initialized to be rounded to the nearest grid, where multiple points may be warped to the same grid. 
% as shown in Fig~\ref{fig:MNN}. 
These points exhibit highly similar features, making it difficult to distinguish them by correlation. 
As a result, supervising each point independently, as in $\mathcal{L}_{r}$, fails to eliminate the quantization error introduced during the initialization stage and leads to over-aggregation issue in the final dense flow.

To this end, we propose a novel loss for refinement, termed flow gradient loss. We not only supervise the values, but also the gradients of $\textbf{f}$ to ensure the continuity of the flow. The flow gradient loss is defined as:
\begin{equation}
\begin{aligned}
    \mathcal{L}_{g} &= \sum_{l}\sum_{u,v} \ \left\|\nabla \textbf{f}_{l}(u,v)_{x} - \nabla \textbf{f}_{l}^{gt}(u,v)_{x}\right\|_{2} \\ &+  \sum_{l}\sum_{u,v}\left\|\nabla \textbf{f}_{l}(u,v)_{y} - \nabla \textbf{f}_{l}^{gt}(u,v)_{y}\right\|_{2},
\end{aligned}    
\end{equation}
where we approximate the gradient $\nabla \textbf{f}$ using finite differences, for example:
\begin{equation}
    \begin{aligned}
    \nabla \textbf{f}_{l}(u,v)_{x} = & \big[\textbf{f}_{l}(u+1,v)_{x} - \textbf{f}_{l}(u,v)_{x},\\& \textbf{f}_{l}(u,v+1)_{x} - \textbf{f}_{l}(u,v)_{x} \big].
    \end{aligned}
\end{equation}

Intuitively, the flow gradient loss encourages the network to learn the pattern of the flow field and the relationships between neighboring points, improving the local consistency of the dense flow significantly.

Overall, the total loss for fine-level stage is $\mathcal{L}_{f}=\mathcal{L}_{r}+\mathcal{L}_{g}$.

\begin{table}[h]
    \caption{\textbf{Number of ground truth matches}
    }
    \centering
    \resizebox{0.9\columnwidth}{!}{
    \setlength\tabcolsep{6pt} %
    \renewcommand{\arraystretch}{1.3}
    \begin{tabular}{lcc} 
    \toprule
    \multirow{2}{*}{Definition}         & \multicolumn{2}{c}{Number of Matches}\\ 
    \cmidrule(lr){2-3}
        &  Sacre Coeur      & St. Peter's Square \\ 
    \midrule
    Based on MNN & 1357.96 & 1301.22  \\
    Ours & 2086.15 & 2137.15 \\ 
    \bottomrule
    \end{tabular}
    }
    \label{tab:num_matches}
\end{table}

\section{Experiments}
\label{experiments}
This section presents extensive experiments on multiple benchmarks, including relative pose estimation, homography estimation and visual localization.
Moreover, various ablation studies are included to provide a comprehensive understanding of our method.

\begin{table*}[t]
    \caption{\textbf{Results of Relative Pose Estimation on MegaDepth Dataset and ScanNet Dataset.} Following~\cite{wang2024efficient}, we evaluate all methods on both datasets using models trained on the MegaDepth dataset, except for Mast3R, which is a fundamental model trained on massive 3D data. The AUC of pose error at different thresholds, along with the processing time for matching image pair at a resolution of 640 × 480, is presented.}
    \centering
    \resizebox{\textwidth}{!}{
    \setlength\tabcolsep{4pt} %
    \renewcommand{\arraystretch}{1.3}
    \begin{tabular}{ccccccccc} 
    \toprule
    \multirow{2}{*}{Category} & \multirow{2}{*}{Method}         & \multicolumn{3}{c}{MegaDepth Dataset}             & \multicolumn{3}{c}{ScanNet Dataset} & \multirow{2}{*}{Time~(ms)} \\ 
    \cmidrule(lr){3-5}
    \cmidrule(lr){6-8}
        &              & AUC@5$\degree$       & AUC@10$\degree$       & AUC@20$\degree$  & AUC@5$\degree$       & AUC@10$\degree$       & AUC@20$\degree$  & \\ 
    \midrule
    \multirow{4}{*}{Sparse} &   SP~\cite{DeTone2017SuperPointSI} + NN & 31.7&46.8&60.1 & 7.5 & 18.6 & 32.1 & 10.8\\
    & SP + SG~\cite{sarlin20superglue} & 49.7& \textbf{67.1} & \textbf{80.6} &  
    16.2 & 32.8 & 49.7 & 48.3 \\
    & SP + LG~\cite{lindenberger2023lightglue} & \textbf{49.9}&67.0&80.1 & 14.8&30.8&47.5&31.9 \\
    & Mast3R~\cite{Leroy2024GroundingIM} & 42.4 & 61.5 & 76.9 & \textbf{28.0} & \textbf{50.2} & \textbf{68.8} & 294.0 \\
    \hline
    &&&&&&&& \\[-3ex]
    \multirow{2}{*}{Dense} & DKM~\cite{edstedt2023dkm} & 60.4 & 74.9 & 85.1 & 26.64 & 47.07 & 64.17 & 210.8\\ %
    & ROMA~\cite{edstedt2023roma} & \textbf{62.6}&\textbf{76.7}&\textbf{86.3} &\textbf{28.9}&\textbf{50.4}&\textbf{68.3}& 302.7\\
    \hline
    &&&&&&&& \\[-3ex]
    \multirow{11}{*}{Semi-Dense}
    & DRC-Net~\cite{li20dualrc} & 27.0&42.9&58.3 &7.7&17.9&30.5 & 328.0 \\
    & LoFTR~\cite{sun2021loftr} & 52.8&69.2&81.2 &16.9&33.6&50.6& 66.2\\
    & QuadTree~\cite{tang2022quadtree} & 54.6&70.5&82.2 &19.0&37.3&53.5 & 100.7\\
    & TopicFM~\cite{giang2022topicfm} & 54.1&70.1&81.6 & 17.3&35.5&50.9 & 66.4 \\
    & AspanFormer~\cite{chen2022aspanformer} & 55.3& 71.5& 83.1 & 19.6 & 37.7 & 54.4 &81.6  \\
    & AdaMatcher~\cite{huang2023adaptive} & 53.5 & 70.3 & 82.6 & 18.8 & 35.0 &50.8& 108.0 \\
    & PATS~\cite{Ni2023PATS} & \cellcolor{tabsecond}58.0 & \cellcolor{tabsecond}73.3 & \cellcolor{tabsecond}84.2 & \cellcolor{tabthird}21.4 & \cellcolor{tabthird}39.8 & \cellcolor{tabsecond}56.8 & 873.7\\
    & EfficientLoFTR~\cite{wang2024efficient} & 56.4& 72.2& 83.5& 19.2&37.0&53.6 & \cellcolor{tabsecond}40.1\\
    & JamMa~\cite{lu2025jamma} & \cellcolor{tabthird}56.8 &  72.6 &  \cellcolor{tabthird}83.6 &  16.4 & 33.1 & 49.9 & \cellcolor{tabfirst}35.8 \\
    & \modify{CoMatch~\cite{li2025comatch}} & \cellcolor{tabsecond}\modify{58.0} & \cellcolor{tabthird}\modify{73.2} & \cellcolor{tabsecond}\modify{84.2} & \cellcolor{tabsecond}\modify{21.7} & \cellcolor{tabfirst}\modify{40.2} & \cellcolor{tabthird}\modify{56.7} & \modify{65.5} \\
    & Ours & \cellcolor{tabfirst}59.6 & \cellcolor{tabfirst}74.3 & \cellcolor{tabfirst}84.7&  \cellcolor{tabfirst}22.0& \cellcolor{tabsecond}40.1& \cellcolor{tabfirst}56.9& \cellcolor{tabthird}56.1\\
    \bottomrule
    \end{tabular}
    }
    \label{tab:exp relativepose}
    \end{table*}

\subsection{Implementation Details}
\paragraph{Feature Extraction and Transformation}
Following~\cite{wang2024efficient}, we adopted RepVGG~\cite{ding2021repvgg} to extract multi-level features and transform coarse features by efficient attention mechanism.
The feature extraction consists of four stages. We use a width of 64 and a stride of 1 for the first stage and widths of [64, 128, 256] and strides of 2 for the subsequent three stages. Each stage is composed of [1, 2, 4, 14] RepVGG blocks with ReLU activations, respectively. The output from the final stage, at $\nicefrac{1}{8}$ of the image resolution, serves as the input for efficient local feature transformer modules to generate attended coarse feature maps. Efficient local feature transformer modules are adopted to generate attended coarse feature maps, with the coarsest local feature as input. 
Each module aggregates query tokens $f_{i}$ using a strided depthwise convolution and key tokens $f_{j}$ using max pooling, both with a kernel size of $ s \times s $. This reduces the token count by $ s^2 $. Positional encoding (PE) with RoPE~\cite{Su2021RoFormerET} is applied only for self-attention, followed by vanilla cross attention. The transformed feature map is then upsampled and fused with $f_{i}$ to produce the final output. This design improves efficiency by reducing redundant computations.

\paragraph{Cascaded Flow Refinement}
Following~\cite{edstedt2023dkm}, cascaded convolutional modules were adopted to refine the flow. 
Each stage refiner predicts a residual offset for the estimated flow from the previous stage, which is bilinearly upsampled to match the scale of the feature maps. They take all channels of $F^{S}$ and the warped $F^{T}$, as well as local correlation in the target view as input via simple concatenation. The refiner blocks use a DenseNet~\cite{Huang_2017_CVPR} architecture with 5x5 depthwise separable kernels, followed by a 1x1 convolution.
The same balanced sampling strategy as~\cite{edstedt2023dkm} was adopted to obtain a proper number of matches for estimating pose or homography from dense matching results.

\paragraph{Supervision}
Our model was trained on the large scale outdoor dataset MegaDepth~\cite{li2018megadepth}, which consists of 196 3D scenes reconstructed from 1M Internet images. To retain all reasonable matches as much as possible using the relative camera pose and depth map, a different ground truth definition from LoFTR~\cite{sun2021loftr} was adopted, which is based on the MNN criterion. According to a cycle-consistency principle, a patch is considered matchable as long as its center point can be warped back to itself after being projected twice. Formally, the ground truth is defined as:
\begin{equation*}
    \mathcal{M}_{c}^{gt}=\{(i, j)| G(F(p_{i}^{A})) \in P_{i}^{A}, F(p_{i}^{A}) \in P_{j}^{B} \},
\end{equation*}
where $F$ and $G$ denote the ground truth warp between the source image and the target image respectively. $P_{i}^{A}$ and $P_{j}^{B}$ are the patches centered at $p_{i}^{A}$ and $p_{j}^{B}$ in the source and target images, respectively.
We compared the number of ground truth matches based on the MNN criterion and the new definition for the scenes of “Sacre Coeur” and “St. Peter's Square” in MegaDepth, as shown in Table~\ref{tab:num_matches}. 

\begin{figure*}[!t]
    \centering
    \includegraphics[width=0.9\linewidth]{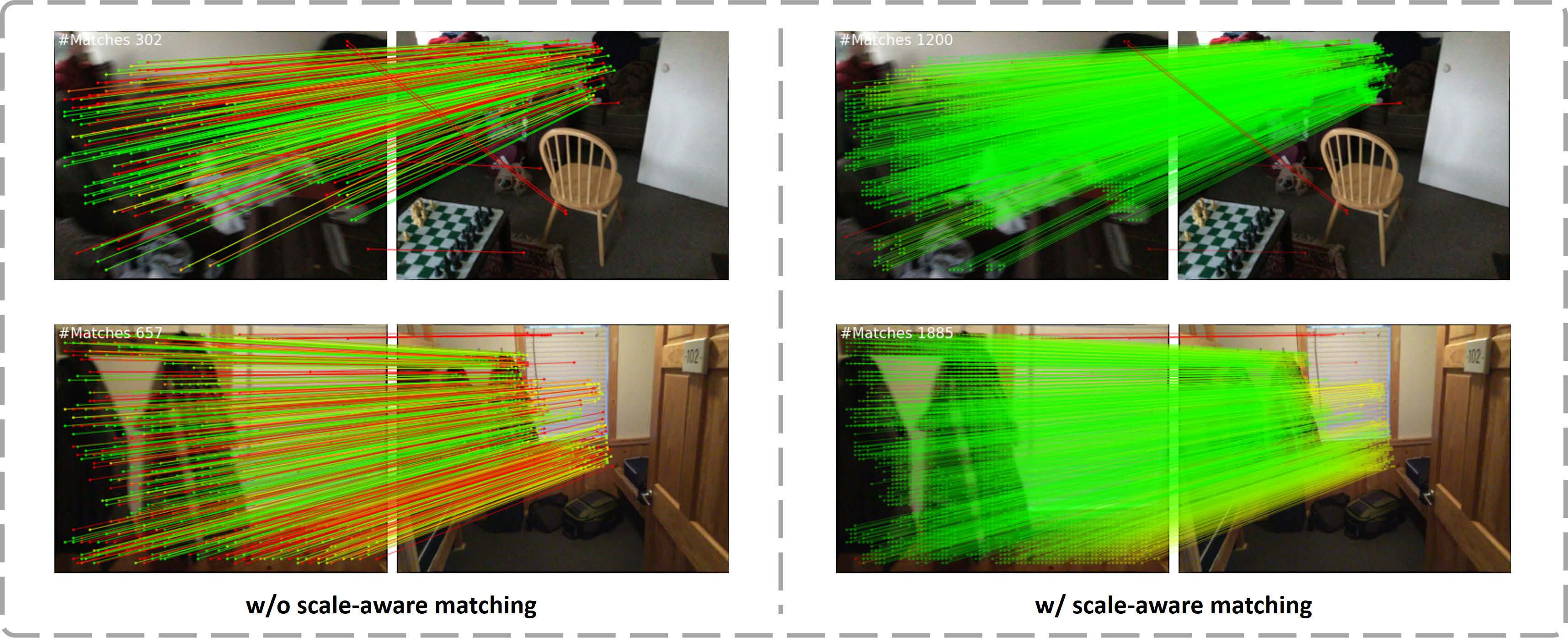}%
    \caption{\textbf{Qualitative comparison results indoor.} Not only the number of the matches but also the matching accuracy demonstrate the superiority of the proposed scale-aware matching mechanism.}
    \label{fig:qualitative_2}
\end{figure*}

\paragraph{Training Details}
The coarse matching module and the cascaded refinement module were trained together from scratch, using the AdamW optimizer with initial learning rates of 4e-3 and 2e-3, respectively.
The training was conducted on 8 NVIDIA 3090 GPUs with batch size of 16, and took about 30 hours to converge. Following~\cite{wang2024efficient},
the trained model on MegaDepth was used to evaluate all datasets and tasks in our experiments to demonstrate the generalization ability.

\subsection{Relative Pose Estimation}
\paragraph{Datasets.}
MegaDepth~\cite{li2018megadepth} dataset and indoor ScanNet~\cite{dai2017scannet} dataset were used for the evaluation of relative pose estimation to demonstrate the effectiveness of the proposed matching pipeline.

MegaDepth dataset is a large-scale outdoor benchmark containing sparse 3D reconstructions, whose camera poses and depth maps are initially computed from COLMAP~\cite{schonbergerStructurefromMotionRevisited2016} and then refined as ground-truth. Following previous work~\cite{wang2024efficient}, relative pose estimation was performed on 1.5k testing pairs, sampled from the scenes of ``Sacre Coeur'' and ``St. Peter's Square'', for fair comparison. \modify{Similar to prior semi-dense evaluations}, images were resized so that the longest edge equals 1200 for all semi-dense and dense methods, while sparse methods were provided resized images with the longest edge equal to 1600. The key challenges of MegaDepth are large viewpoint changes and repetitive structures.

ScanNet dataset is composed of indoor monocular sequences with ground truth poses and depth maps. Wide baselines and extensive textureless regions in image pairs are the key challenges of ScanNet dataset. We followed the test split of~\cite{sarlin20superglue} for the evaluation, and images were resized to $640\times480$ for all methods.

\paragraph{Evaluation protocol.}
Following previous methods, relative poses are recovered by solving the essential matrix from the produced correspondences with RANSAC. We report the cumulative curve (AUC) of the pose error at thresholds (5$^\circ$, 10$^\circ$, and 20$^\circ$), where the pose error is defined as the maximum of angular error in rotation and translation. Moreover, the efficiency of all methods was also evaluated by testing the runtime for matching each image pair in the ScanNet dataset using a single NVIDIA 3090, to further demonstrate the trade-off between matching accuracy and efficiency.

\paragraph{Baseline.}
We compare the proposed method with three categories of methods: 1) sparse methods, including SuperPoint~\cite{DeTone2017SuperPointSI} with Nearest-Neighbor~(NN), SuperGlue~(SG)~\cite{sarlin20superglue}, LightGlue~(LG)~\cite{lindenberger2023lightglue}, Mast3R~\cite{Leroy2024GroundingIM} , 2) semi-dense matcher, including DRC-Net~\cite{li20dualrc}, LoFTR~\cite{sun2021loftr}, QuadTree Attention~\cite{tang2022quadtree}, AspanFormer~\cite{chen2022aspanformer}, TopicFM~\cite{giang2022topicfm}, AdaMatcher~\cite{huang2023adaptive}, PATS~\cite{Ni2023PATS}, EfficientLoFTR~\cite{wang2024efficient}, JamMa~\cite{lu2025jamma}\modify{, CoMatch~\cite{li2025comatch}} and 3) dense matcher, including DKM~\cite{edstedt2023dkm}, ROMA~\cite{edstedt2023roma}.

\paragraph{Results.}
As shown in Table~\ref{tab:exp relativepose}, our method  significantly benefits from the advantages of the novel matching pipeline. Specifically, compared with EfficientLoFTR~\cite{wang2024efficient} which shares the same feature extraction and transformation stage with us, our method achieves significant improvements on both datasets.
Moreover, compared with the two other methods~\cite{huang2023adaptive, Ni2023PATS} designed to address the over-exclusion issue, our method features a concise and efficient coarse matching stage, evidenced by the inference time.
Although the dense matcher DKM~\cite{edstedt2023dkm} and ROMA~\cite{edstedt2023roma} achieve better performance, our method is $\sim4$ and $\sim5.5$ times faster than them, respectively, which is crucial for real-time applications.
As for Mast3R~\cite{Leroy2024GroundingIM}, it performs well on ScanNet, which is attributed to the fact that it is a fundamental model trained on massive 3D reconstruction data, while other methods are only trained on MegaDepth.
\modify{We also compare with CoMatch~\cite{li2025comatch}, a recent semi-dense method featuring state-of-the-art performance. Our method surpasses CoMatch on MegaDepth while maintaining competitive performance on ScanNet.}

\begin{table}[h]
    \caption{\textbf{Results of Homography Estimation on HPatches Dataset.}
    }
    \centering
    \resizebox{1.0\columnwidth}{!}{
    \setlength\tabcolsep{10.7pt} %
    \renewcommand{\arraystretch}{1.3}
    \begin{tabular}{cccc} 
    \toprule
    \multirow{2}{*}{Method}         & \multicolumn{3}{c}{Homography est. AUC} \\ 
    \cmidrule(lr){2-4}         & @3px       & @5px       & @10px \\ 
    \midrule
    SP + SG~\cite{sarlin20superglue} & 53.9 & 68.3 & 81.7 \\
    Sparse-NCNet~\cite{rocco2020efficient} & 48.9 & 54.2 & 67.1 \\
    DRC-Net~\cite{li20dualrc} & 50.6 & 56.2 & 68.3 \\
    LoFTR~\cite{sun2021loftr} & 65.9 & 75.6 & 84.6\\
    AspanFormer~\cite{chen2022aspanformer} & 67.4 & 76.9 & 85.4 \\
    EfficientLoFTR~\cite{wang2024efficient} & 66.5 & 76.4 & \cellcolor{tabthird}85.5 \\
    JamMa~\cite{lu2025jamma} & \cellcolor{tabthird}68.1 & \cellcolor{tabthird}77.0 & 85.4 \\
    \modify{CoMatch~\cite{li2025comatch}} & \cellcolor{tabsecond}\modify{68.4} & \cellcolor{tabsecond}\modify{78.2} & \cellcolor{tabsecond}\modify{86.8} \\
    Ours & \cellcolor{tabfirst}70.5 & \cellcolor{tabfirst}79.6 & \cellcolor{tabfirst}87.1 \\
    \bottomrule
    \end{tabular}
    }
    \label{tab:exp hpatches}
    \end{table}

\subsection{Homography Estimation}
\paragraph{Dataset.}
We used HPatches~\cite{balntas2017hpatches} dataset for the evaluation of homography estimation to demonstrate the efficacy of our method. HPatches contains 52 sequences under significant illumination changes and 56 sequences that exhibit large variation in viewpoints, \modify{which makes it a standard benchmark for planar matching and homography estimation}.

\paragraph{Evaluation Protocol.}
We followed the evaluation protocol proposed by LoFTR~\cite{sun2021loftr}, resizing the smallest edge of the images to 480. We report the area under the AUC of the corner error distance up to 3, 5, and 10 pixels, respectively. The same RANSAC method is employed as a robust homography estimator for all baselines for fair comparison.

\paragraph{Results.}
As shown in Table~\ref{tab:exp hpatches}, our method outperforms the best semi-dense baselines \modify{including CoMatch~\cite{li2025comatch},} AspanFormer~\cite{chen2022aspanformer} and JamMa~\cite{lu2025jamma}. 

\subsection{Visual Localization}
\paragraph{Datasets and Evaluation Protocols.}
\modify{Visual localization is the task of estimating the 6-DoF poses of given images with respect to the corresponding 3D scene model. We evaluated our method on InLoc~\cite{taira2018inloc} dataset and Aachen v1.1~\cite{sattler2018benchmarking} dataset using the open-source localization framework HLoc. InLoc is an indoor benchmark characterized by plenty of texture-less areas and repetitive structures, while Aachen v1.1 is a outdoor benchmark with day-night appearance changes. Following the benchmark settings, we report the percentage of pose errors satisfying both angular and distance thresholds on the corresponding evaluation splits}.

\paragraph{Results.}
As shown in Table~\ref{tab:exp inloc} and Table~\ref{tab:exp aachen}, our method achieves competitive performance on visual localization, which verifies the effectiveness of the novel pipeline. \modify{We also include the result of DKM~\cite{edstedt2023dkm} on Aachen v1.1 for reference. Notably, even this strong dense matcher only achieves 89.7/95.9/99.2 on Day and 77.0/90.1/99.5 on Night, suggesting that the benchmark is already highly saturated and leaves limited room for large numerical gains.}

\begin{table}[ht]
    \caption{\textbf{Results of Visual Localization on InLoc Dataset.}
    }
    \centering
    \resizebox{1.0\columnwidth}{!}{
    \setlength\tabcolsep{4.5pt} %
    \renewcommand{\arraystretch}{1.3}
    \begin{tabular}{ccccccc} 
    \toprule
    \multirow{2}{*}{Method}         &  \multicolumn{3}{c}{DUC1} & \multicolumn{3}{c}{DUC2}\\ 
    \cmidrule(lr){2-7}
        &              \multicolumn{6}{c}{(0.25m,2$\degree$)/(0.5m,5$\degree$)/(1.0m,10$\degree$)} \\ 
    \midrule
    SP+SG~\cite{sarlin20superglue} & 49.0 & 68.7 & 80.8 & 53.4 & 77.1 & 82.4 \\
    DRC-Net~\cite{li20dualrc} & 40.6 & 65.6 & 70.1 & 52.8 & 76.0 & 74.3 \\
    LoFTR~\cite{sun2021loftr} & 47.5 & 72.2 & 84.8 & 54.2 & 74.8 & 85.5 \\
    TopicFM~\cite{giang2022topicfm} & 52.0 & \cellcolor{tabthird}74.7 & \cellcolor{tabsecond}87.4 & 53.4 & 74.8 & 83.2 \\
    PATS~\cite{Ni2023PATS} & \cellcolor{tabfirst}55.6 & 71.2 & 81.0 & \cellcolor{tabthird}58.8 & \cellcolor{tabthird}80.9 & 85.5 \\
    AspanFormer~\cite{chen2022aspanformer} & 51.5 & 73.7 & 86.0 & 55.0 & 74.0 & 81.7\\
    EfficientLoFTR~\cite{wang2024efficient} & 52.0 & \cellcolor{tabthird}74.7 & \cellcolor{tabthird}86.9 & 58.0 & \cellcolor{tabthird}80.9 & \cellcolor{tabfirst}89.3 \\
    \modify{CoMatch~\cite{li2025comatch}} & \cellcolor{tabsecond}\modify{54.5} & \cellcolor{tabsecond}\modify{75.3} & \cellcolor{tabthird}\modify{86.9} & \cellcolor{tabsecond}\modify{59.5} & \cellcolor{tabfirst}\modify{82.1} & \cellcolor{tabthird}\modify{87.8} \\
    Ours & \cellcolor{tabthird}53.7 & \cellcolor{tabfirst}75.5 & \cellcolor{tabfirst}87.5 & \cellcolor{tabfirst}59.9 & \cellcolor{tabsecond}81.8 & \cellcolor{tabsecond}88.5 \\
    \bottomrule
    \end{tabular}
    }
    \label{tab:exp inloc}
    \end{table}

\begin{table}[ht]
    \caption{\textbf{Results of Visual Localization on Aachen v1.1 Dataset.}
    }
    \centering
    \resizebox{1.0\columnwidth}{!}{
    \setlength\tabcolsep{4.5pt} %
    \renewcommand{\arraystretch}{1.3}
    \begin{tabular}{ccccccc} 
    \toprule
    \multirow{2}{*}{Method}         &  \multicolumn{3}{c}{Day} & \multicolumn{3}{c}{Night}\\ 
    \cmidrule(lr){2-7}
        &              \multicolumn{6}{c}{(0.25m,2$\degree$)/(0.5m,5$\degree$)/(1.0m,10$\degree$)} \\ 
    \midrule
    SP+SG~\cite{sarlin20superglue} & \cellcolor{tabsecond}{89.8} & \cellcolor{tabsecond}{96.1} & \cellcolor{tabfirst}{99.4} & 77.0 & 90.6 & \cellcolor{tabfirst}{100.0} \\
    LoFTR~\cite{sun2021loftr} & 88.7 & 95.6 & 99.0 & \cellcolor{tabfirst}{78.5} & 90.6 & 99.0 \\
    TopicFM~\cite{giang2022topicfm} & \cellcolor{tabfirst}{90.2} & \cellcolor{tabthird}{95.9} & 98.9 & \cellcolor{tabthird}{77.5} & \cellcolor{tabthird}{91.1} & \cellcolor{tabthird}{99.5} \\
    PATS~\cite{Ni2023PATS} &  \cellcolor{tabthird}{89.6} & 95.8 & \cellcolor{tabsecond}{99.3} & 73.8 & \cellcolor{tabfirst}{92.1} & \cellcolor{tabthird}{99.5} \\ 
    AspanFormer~\cite{chen2022aspanformer} & 89.4 & 95.6 & 99.0 & \cellcolor{tabthird}{77.5} & \cellcolor{tabsecond}{91.6} & \cellcolor{tabthird}{99.5}\\
    EfficientLoFTR~\cite{wang2024efficient} & \cellcolor{tabthird}{89.6} & \cellcolor{tabfirst}{96.2} & 99.0 & 77.0 & \cellcolor{tabthird}{91.1} & \cellcolor{tabthird}{99.5} \\
    \modify{CoMatch~\cite{li2025comatch}} & \modify{89.4} & \modify{95.8} & \modify{99.0} & \cellcolor{tabfirst}\modify{78.5} & \cellcolor{tabsecond}\modify{91.6} & \cellcolor{tabthird}\modify{99.5} \\
    \modify{DKM~\cite{edstedt2023dkm}} & \modify{89.7} & \modify{95.9} & \modify{99.2} & \modify{77.0} & \modify{90.1} & \modify{99.5} \\
    Ours & \cellcolor{tabsecond}{89.8} & \cellcolor{tabfirst}{96.2} & \cellcolor{tabthird}{99.1} & \cellcolor{tabsecond}{77.9} & 91.0 & \cellcolor{tabsecond}99.6 \\
    \bottomrule
    \end{tabular}
    }
    \label{tab:exp aachen}
    \end{table}

\subsection{Ablation Studies}
\paragraph{Effectiveness of Each Design.}
We first conducted ablation studies to analyze the effectiveness of each design in our method on MegaDepth dataset with longest edge of the image set to 1200, as shown in Table~\ref{tab:ablation}.
(1) Directly performing cascaded flow refinement on the unreliable coarse results of standard MNN matching leads to performance degradation, which even performs worse than the baseline method~\cite{wang2024efficient}.
(2) Compared with the refinement stage of the baseline~\cite{wang2024efficient}, our cascaded flow refinement can bring significant accuracy improvements.
(3) If we discard the flow gradient loss, the local consistency of the flow field will be degraded, which can be reflected in the cumulative curve (AUC) of the pose error.

\begin{table}[ht]
    \caption{\textbf{Effectiveness of Each Design.}}
    \centering
    \resizebox{1\columnwidth}{!}{
    \setlength\tabcolsep{1.5pt} %
    \renewcommand{\arraystretch}{1.3}
    
    \begin{tabular}{lcccc} 
    \toprule
    \multirow{2}{*}{Method}         & \multicolumn{3}{c}{Pose Estimation AUC}\\ 
    \cmidrule(lr){2-4}
        & @5\degree       & @10\degree       & @20\degree \\ 
    \midrule
    Ours Full & 59.6 & 74.3 & 84.7  \\
    (1) w/o scale-aware coarse matching & 56.1 & 71.9 & 83.3 \\
    (2) Replace our refine. with vanilla refine. & 57.4 & 72.9 & 83.5 \\
    (3) w/o flow gradient loss & 58.9 & 73.8 & 84.0 \\ 
    
    \bottomrule
    \end{tabular}
    }
    \label{tab:ablation}
    \end{table}

\begin{figure*}[!ht]
    \centering
    \includegraphics[width=0.9\linewidth]{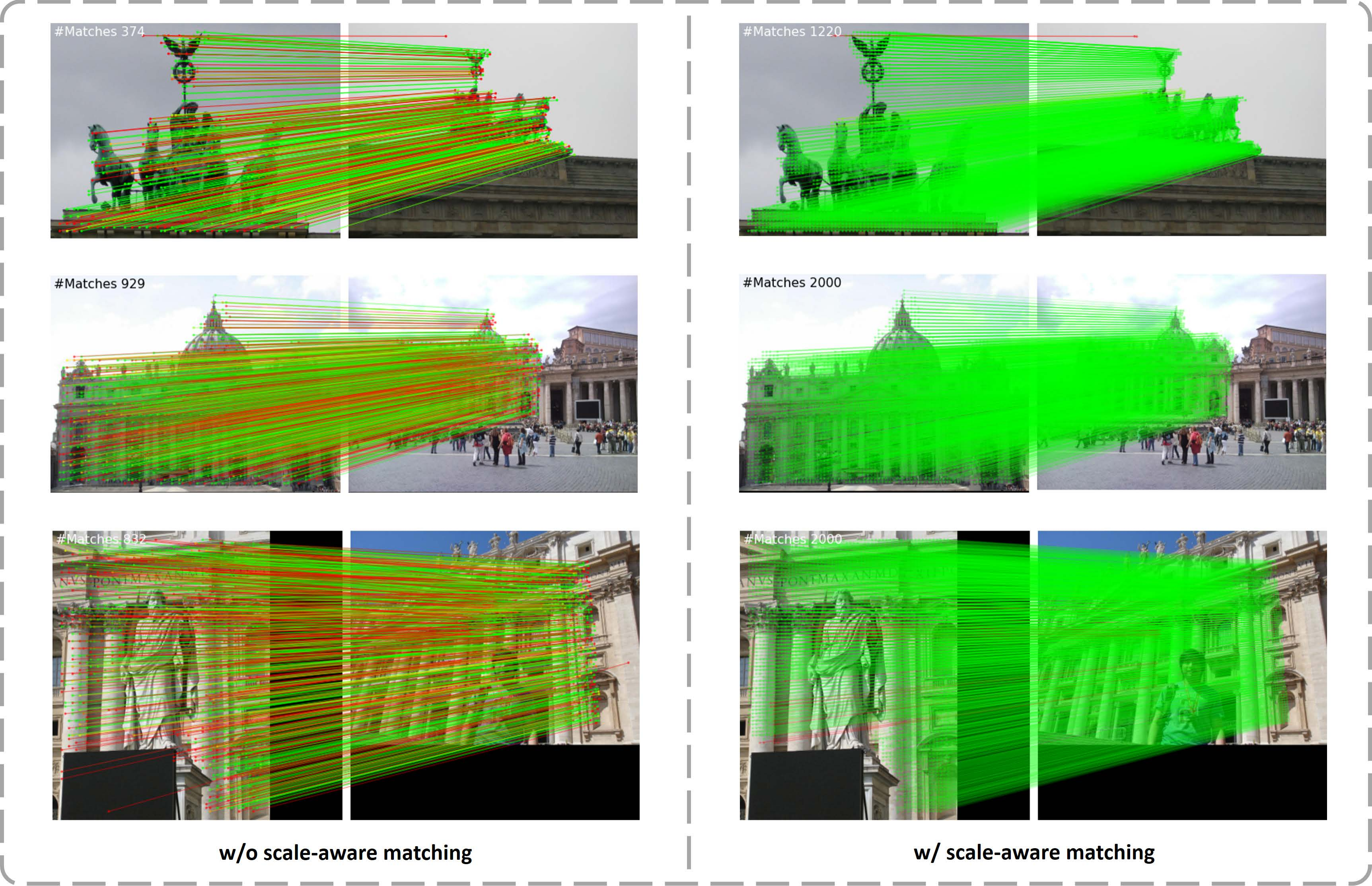}%
    \caption{\textbf{Qualitative comparison results outdoor.} The left and right columns show the matching results w/o and w/ scale-aware matching mechanism respectively. It can be observed that scale-aware matching mechanism not only increases the number of matches but also improves the matching accuracy. }
    \label{fig:qualitative_1}
\end{figure*}

\begin{table}[h]
    \caption{\textbf{AUC 5$^\circ$ vs. Scale ratio.}}
    \centering
    \resizebox{1.0\columnwidth}{!}{
    \setlength\tabcolsep{3pt} %
    \renewcommand{\arraystretch}{1.3}
    \begin{tabular}{ccccccc} 
    \toprule
    \multirow{2}{*}{Method}         &  \multicolumn{5}{c}{Pose Estimation AUC@5$\degree$}\\ 
    \cmidrule(lr){2-6}
        &              $(0, 1)$ & $[1, 2)$ & $[2, 3)$ & $[3, 4)$ & $[4, +\infty)$ \\ 
    \midrule
    AspanFormer~\cite{chen2022aspanformer} & 58.7 & 59.0 & 54.7 & 51.6 & 46.3 \\
    AdaMatcher~\cite{huang2023adaptive} & 58.1 & 58.8 & \cellcolor{tabthird}56.0 & \cellcolor{tabthird}54.8 & \cellcolor{tabthird}52.5 \\ 
    PATS~\cite{Ni2023PATS} & \cellcolor{tabsecond}61.6 & \cellcolor{tabsecond}62.5 & \cellcolor{tabsecond}59.9 & \cellcolor{tabfirst}58.0 & \cellcolor{tabsecond}54.8  \\
     EfficientLoFTR~\cite{wang2024efficient} & \cellcolor{tabthird}61.2 & \cellcolor{tabthird}60.2 & 55.1 & 52.4 & 46.7  \\
    Ours & \cellcolor{tabfirst}62.3 & \cellcolor{tabfirst}63.7 & \cellcolor{tabfirst}60.3 & \cellcolor{tabfirst}58.0 & \cellcolor{tabfirst}55.1 \\
    \bottomrule
    \end{tabular}
    }
    \label{tab:scale}
\end{table}

\paragraph{Scale Variance}
Scale variance between images is a challenge that image matching methods suffer from. To verify the robustness of our method, we use a scale-split MegaDepth test set, sampled from 6 scenes, to evaluate the impact of scale variance on relative pose estimation. The scale ratio ranges in $(0, 1)$, $[1, 2)$, $[2, 3)$, $[3, 4)$ and $[4, +\infty)$, where the scale ratio is defined as proportion of co-visible patches between the source and target images, and obtained based on the ground truth.
We choose EfficientLoFTR~\cite{wang2024efficient} which shares the same feature processing with us as baseline, as well as AdaMatcher~\cite{huang2023adaptive} and PATS~\cite{Ni2023PATS} which are also designed for addressing the over exclusion issue of MNN matching. As shown in Table~\ref{tab:scale}, the superiority of our method over EfficientLoFTR becomes increasingly evident as the scale ratio increases. This is attributed to our flexible matching module, which effectively handles many-to-one matching scenarios. More qualitative results can be seen in Fig.~\ref{fig:qualitative_1} and Fig.~\ref{fig:qualitative_2}.

\begin{figure}[ht!]
    \centering
    \includegraphics[width=0.8\linewidth]{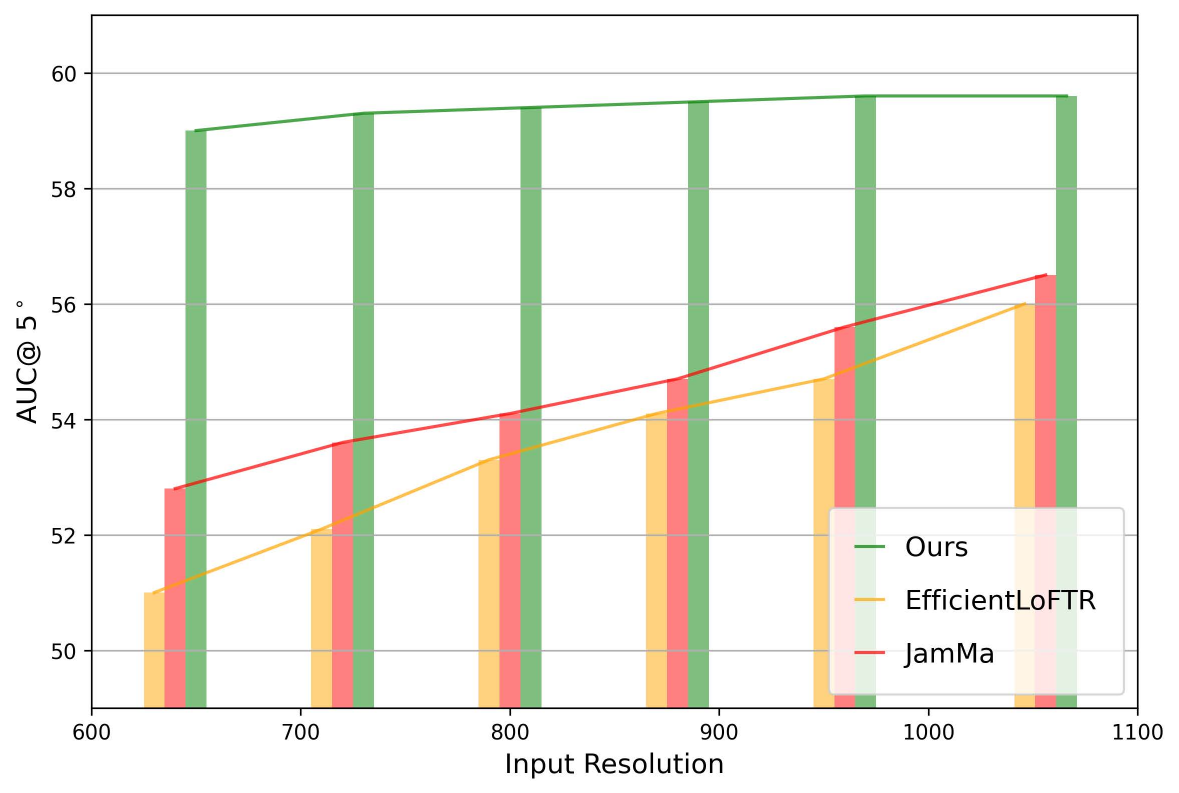}
    \caption{\textbf{AUC@5$^\circ$ vs. Resolution.} Our Method is robust against the variation in image resolution, compared with EfficientLoFTR~\cite{wang2024efficient} and JamMa~\cite{lu2025jamma}. Notably, when the resolution is 640, our method outperforms the baselines by 9.0 and 6.1 on AUC@5$^\circ$, respectively.}
    \label{fig:resolution}%
\end{figure}

\paragraph{Image Resolution}
We conducted experiments on relative pose estimation using MegaDepth dataset to verify the robustness of our method against variations in image resolution.
We choose EfficientLoFTR~\cite{wang2024efficient} as the baseline method for comparison, which shares the same feature extraction and transformation stage with us.
In our experiment, we adjust the longest edge of the images to 640, 736, 832, 960, and 1056 pixels, and report the AUC of the pose error at a threshold of 5$^\circ$.
As demonstrated in Fig.~\ref{fig:resolution}, the pose estimation accuracy of our method remains barely affected by changes in image resolution, while the performance of EfficientLoFTR shows a significant decline as the resolution decreases.
\vspace{0.5cm}

\paragraph{Details About Timing}
Following the same setting as~\cite{wang2024efficient}, we evaluate the inference time of our method on a single NVIDIA 3090 GPU. The running times evaluated in the paper are averaged over all pairs in the test dataset with a warm-up of 50 pairs for accurate measurement. We further report each part running time of our method in Table~\ref{tab: timing}.

\begin{table}[h]
    \caption{ 
    \textbf{Time cost for an image pair of $640\times480$.}
    }
    \centering
    \resizebox{0.9\columnwidth}{!}{
    \setlength\tabcolsep{13pt} %
    \renewcommand{\arraystretch}{1.5}
    \begin{tabular}{lc} 
    \toprule
    Process        & Time(ms)  \\ %
    \midrule
    Total & 56.1 \\
    \hline
    Feature Backbone & 9.1 \\
    Coarse Feature Transformation & 11.7  \\
    Coarse Matching & 8.7 \\
    Cascaded Flow Refinement & 26.6 \\
    \bottomrule
    \end{tabular}
    }
    \label{tab: timing}
\end{table}

% \begin{color}{red}
\section{Discussion on Our Method}
\label{sec:discussion}

\begin{figure}[ht]
    \centering
    \includegraphics[width=\columnwidth]{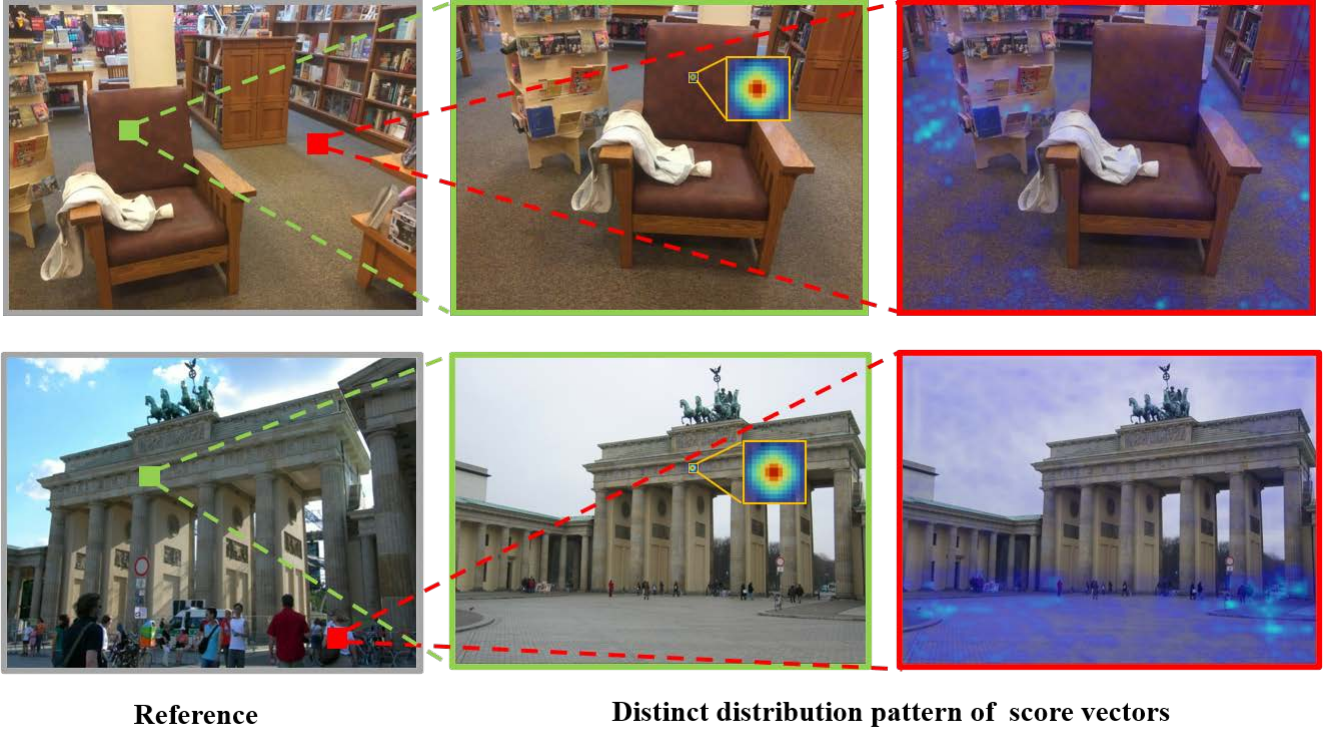}
    \vspace{-0.8cm}
    \caption{\textbf{Qualitative analysis in repetitive-texture scene.} We compare representative matching-score heatmaps of co-visible repetitive points and unmatchable points. The co-visible repetitive points still show locally concentrated responses, while the unmatchable point exhibits broadly diffused responses.}
    \label{fig:qualitative_repetitive}
\end{figure}

\paragraph{Entropy Discriminability in Repetitive Regions.}
A potential concern is that both co-visible repetitive regions and unmatchable points may produce high entropy, which would reduce the separability of a single entropy threshold. To analyze these cases, we provide dedicated qualitative examples in Fig.~\ref{fig:qualitative_repetitive}. In these repetitive-texture pairs, we compare the matching-score heatmaps of a co-visible repetitive point and an unmatchable point. 
The response of the co-visible repetitive point remains locally concentrated around the correct correspondence. By contrast, the unmatchable point yields a much more broadly diffused response over a large area. Therefore, the entropy increase caused by repetitive texture is limited and remains clearly lower than that of truly unmatchable points in practice, so entropy-threshold-based separation remains effective.

This behavior is consistent with the coarse-stage design inherited from LoFTR. The transformer-based global receptive field and positional encoding provide contextual and spatial priors, improving discriminability in repetitive regions and reducing full-map ambiguity. Consequently, entropy remains a useful cue for co-visibility estimation.

\paragraph{Threshold Selection and Sensitivity.}
The threshold $\theta_e$ separates low-entropy co-visible points from high-entropy unmatchable points. We conducted an ablation study on scale-balance-sampled MegaDepth (outdoor) and ScanNet (indoor) to identify appropriate default values, as shown in Table~\ref{tab:ablation_threshold}. Based on these results, we set $\theta_e=3$ for outdoor and $\theta_e=4$ for indoor by default. The optimal threshold value for indoor benchmarks is marginally higher, as the ScanNet dataset contains more repetitive-texture scenes which shift the entropy distribution of co-visible patches upward.

\begin{table}[h]
    \caption{\textbf{Ablation of entropy threshold $\theta_e$}}
    \centering
    \resizebox{0.85\columnwidth}{!}{
    \setlength\tabcolsep{11pt} %
    \renewcommand{\arraystretch}{1.2}

    \begin{tabular}{ccccc}
    \toprule
    \multirow{2}{*}{$\theta_e$} & \multicolumn{2}{c}{Outdoor} & \multicolumn{2}{c}{Indoor} \\
    \cmidrule(lr){2-5}
     & \multicolumn{4}{c}{$\left|\textstyle\frac{\hat{\mathcal{R}}-\mathcal{R}}{\mathcal{R}}\right|$ / AUC@5$\degree$} \\
     \midrule
    /&/&51.0 & / & 20.5  \\
    1&0.38&51.4 & 0.42 & 20.0 \\
    2&0.23&53.3 & 0.30 & 20.9\\
    3&\cellcolor{tabfirst}0.09&\cellcolor{tabfirst}55.1 & 0.21 & 21.4\\
    4&0.13&55.0 & \cellcolor{tabfirst}0.14 & \cellcolor{tabfirst}21.8 \\
    5&0.24&53.4 & 0.19 & 21.6 \\

    \bottomrule
    \end{tabular}
    }
    \label{tab:ablation_threshold}
\end{table}

Although the optimal thresholds differ between domains, the performance is not highly sensitive to this hyperparameter. For instance, using a unified threshold of $\theta_e=4$ for both indoor and outdoor data results in minimal performance degradation on outdoor datasets (AUC@5$^\circ$: 55.1 $\to$ 55.0), while both settings substantially outperform the baseline without AMNN (51.0). This indicates that reasonable threshold choices yield consistent improvements across domains.

In practice, a simple protocol suffices: tune $\theta_e$ on the training/validation split of the target benchmark family and apply the selected value to testing. This approach is stable under commonly used benchmark settings where train and test domains are not extremely mismatched.

\paragraph{Global Nature of Scale Approximation.}
Our scale ratio $\mathcal{R}=|\mathcal{V}_s|/|\mathcal{V}_t|$ should be understood as a coarse global approximation of the dominant scale tendency between the two images, rather than an explicit model of spatially varying local scale. In our method, this ratio is only used to determine the inspection window size in AMNN, rather than to directly predict the final correspondences. Therefore, its practical role is to provide a reasonable bias for relaxing the overly strict one-to-one constraint of standard MNN. We agree that under extreme non-uniform scale changes caused by severe perspective distortion or large depth discontinuities, the estimated ratio may become less accurate. Nevertheless, the results in Table~\ref{tab:ablation_threshold} suggest that the practical robustness of our method does not rely on exact scale recovery. For example, on the outdoor benchmark, using $\theta_e=4$ yields a slightly less accurate scale estimate than $\theta_e=3$ (0.13 vs. 0.09), but the AUC@5$^\circ$ remains nearly unchanged (55.0 vs. 55.1) and still clearly outperforms the baseline without AMNN (51.0). This evidence suggests that the estimated ratio mainly serves as a sufficiently reliable global cue for setting the AMNN inspection window, which is already enough to improve coarse matching robustness in practice.

\paragraph{Comparison with Dense Matching Methods.}
Although our method significantly improves the robustness of semi-dense matching by reinforcing the coarse matching stage against scale variance, it remains less robust than dense matchers like DKM in highly challenging cases. This remaining gap lies primarily in the coarse matching stage, specifically under extreme viewpoint or appearance changes that fall outside the scope of this work.
Specifically, DKM starts from a dense warp prediction at a very coarse resolution (1/32) and progressively refines it, ensuring reliable initialization at the cost of substantially higher computation.
In contrast, semi-dense methods, including ours, establish coarse correspondences directly via a differential matching layer with negligible overhead after 1/8 feature extraction.
While this design prioritizes efficiency, it is more vulnerable to extreme viewpoint or appearance changes compared to dense methods.
Consequently, once coarse correspondences are inaccurate, the subsequent cascaded flow refinement can only optimize them locally rather than recovering a globally reliable field from scratch.
% \end{color}

\vspace{-0.2cm}
\section{Conclusion and Limitations}
\label{sec:conclusion}
This paper comprehensively revisits the existing semi-dense matching pipeline to improve the robustness and accuracy of local feature matching.
Specifically, we address two long-standing issues in the coarse and fine stages, respectively, through scale-aware matching and cascaded flow refinement.
Extensive experiments demonstrate that our method achieves superior performance on downstream tasks and performs robustly against image resolution and scale variation, compared to existing semi-dense methods.

\modify{
However, our method still has several limitations. First, although our scale-aware design improves the robustness of coarse matching under practical scale variation, this approximation may become less reliable under strongly non-uniform scale changes, such as severe perspective distortion, where the local scale may vary significantly across the image. Second, while our method substantially strengthens semi-dense matching, it is still slightly inferior to dense methods~\cite{edstedt2023dkm, edstedt2023roma} on downstream tasks. In particular, This remaining gap lies primarily in the coarse matching stage, specifically under extreme viewpoint or appearance changes that fall outside the scope of this work. In such situations, dense methods benefit from progressively refining a dense warp prediction from a very coarse resolution (1/32), ensuring reliable coarse matching at the cost of substantially higher computation. Furthermore, whether our architecture can scale to massive 3D data as in~\cite{Leroy2024GroundingIM} remains an open question. We leave these issues for future work.}

\ifCLASSOPTIONcaptionsoff
  \newpage
\fi

\newpage
\bibliographystyle{IEEEtran}
\bibliography{ref}

% \newpage
% \vspace{5cm}
\begin{IEEEbiography}
[{\includegraphics[width=1in,height=1.25in,clip,keepaspectratio]{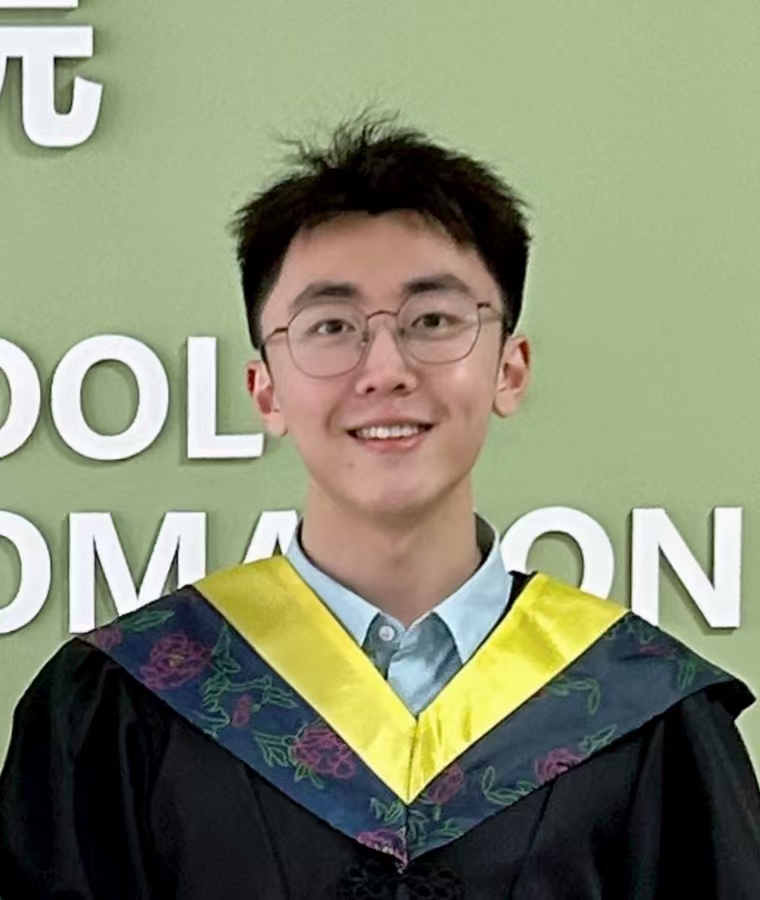}}]{Ke Jin} received the B.S. degree from the College of Automation at Southeast University, Nanjing, China, in 2024. He is currently pursuing the Ph.D. degree with the College of Control Science and Engineering at Zhejiang University, Hangzhou, China. His research interests lie in image matching, pose estimation and dexterous manipulation.
\end{IEEEbiography}

% \vspace{-7cm}
\begin{IEEEbiography}
[{\includegraphics[width=1in,height=1.25in,clip,keepaspectratio]{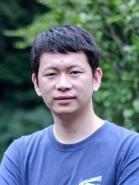}}]{Jiming Chen} is a Changjiang Scholars Professor with College of control science and engineering, Zhejiang University. He is vice Dean of Facutly of Information Technology,  deputy director of the State Key laboratory of Industrial Control Technology, Director of Industrial Process Control. He was a visiting researcher at University of Waterloo from 2008 to 2010. Currently, He serves/served associate editors for ACM TECS, IEEE TPDS, IEEE Network, IEEE TCNS, IEEE TII, etc. He has been appointed as a distinguished lecturer of IEEE Vehicular Technology Society 2015, and selected in National Program for Special Support of Top-Notch Young Professionals, and also funded Excellent Youth Foundation of NSFC. He also was the recipients of IEEE INFOCOME 2014 Best Demo Award, IEEE ICCC 2014 best paper award, IEEE PIMRC 2012 best paper award, and JSPS Visiting Fellowship 2011. He also received the IEEE Comsoc Asia-pacific Outstanding Young Researcher Award 2011. He is a Distinguished Lecturer of IEEE Vehicular Technology Society (2015-2018), and a Felllow of IEEE. His research interests include networked control, sensor networks, cyber security, IoT and autonomous intelligent unmanned system.
\end{IEEEbiography}

% \vspace{-7cm}
\begin{IEEEbiography}
[{\includegraphics[width=1in,height=1.25in,clip,keepaspectratio]{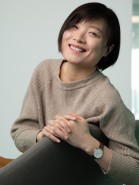}}]{Qi Ye} is a Tenure-Track Professor under the Hundred Talents Program at Zhejiang University. Before joining Zhejiang University in 2020, She was a research scientist of Mixed Reality \& AI Lab at Cambridge, Microsoft. Prior to joining Microsoft, she was a Ph.D. student at Imperial Computer Vision and Learning Lab of Imperial College London, under the supervision of Tae-Kyun Kim. She obtained her Master's degree from the School of Information Science and Technology of Tsinghua University in 2014 and a Bachelor's degree from the College of Information Science and Technology of Beijing Normal University in 2011. She is interested in and working on human-computer interaction, particularly vision understanding involving hands, and a more general setting where humans interact with the environment. She is passionate about 3D vision and their applications, particularly in Mixed/Augmented/Virtual Reality and automatic control.
\end{IEEEbiography}

\end{document}